\RequirePackage{fix-cm}
\documentclass[smallextended]{svjour3}       
\smartqed  
\usepackage{graphicx}
\usepackage{epsfig}
\usepackage{amsmath}
\usepackage{amssymb}
\usepackage{url}
\usepackage{multirow}
\usepackage{array}
\begin{document}

\title{Towards a Solid Solution of Real-time Fire and Flame Detection
}


\author{Bo JIANG \and Yongyi LU \and Xiying LI \and Liang LIN  
}


\institute{Bo JIANG \at
              Sun Yat-sen University,Guangzhou,China \\
              \email{jiangbo3@mail.sysu.edu.cn}           
           \and
           Yongyi LU \at
             Sun Yat-sen University,Guangzhou,China
           \and
           Xiying LI \at
             Sun Yat-sen University,Guangzhou,China
           \and
           Liang LIN \at
             Sun Yat-sen University,Guangzhou,China
}


\maketitle

\begin{abstract}
Although the object detection and recognition has received growing attention for decades, a robust fire and flame detection method is rarely explored.
This paper presents an empirical study, towards a general and solid approach to fast detect fire and flame in videos, with the applications in video surveillance and event retrieval. Our system consists of three cascaded steps: (1) candidate regions proposing by a background model, (2) fire region classifying with color-texture features and a
dictionary of visual words, and (3) temporal verifying. The experimental evaluation and analysis are done for each step. We believe that it is a useful service
to both academic research and real-world application. In addition, we release the software of the proposed system with the source code, as well as a public benchmark and data set, including 64 video clips covered both indoor and outdoor scenes under different conditions. We achieve an 82\% Recall with 93\% Precision on the data set, and greatly improve the performance by state-of-the-arts methods.
\keywords{Fire detection \and Empirical study \and Video surveillance \and Region classification}
\end{abstract}

\section{Introduction}
\label{sec:intro}
Fire catastrophes always cause big lost to human kind. For example, Dhaka fire on June 2010 caused over 117 people died and over 100 injured. Forest fires in Russia in summer 2010 caused over 10 billion US dollars lost. Therefore, robust early-warning fire alarm systems are always in need to protect people's safety and properties. An fast and accurate fire detection method is the key technology needed in such systems. As a result, fire or flame detection has received great attention by research community for decades~\cite{Toreyin07CVPR}\cite{Ko09FireSVM}. In recent years, fire detection also find applications in event retrieval on fire in digital image or video archives~\cite{PVKBorges10}.
\begin{figure}[htb]
\begin{center}
\includegraphics[width=0.98\linewidth]{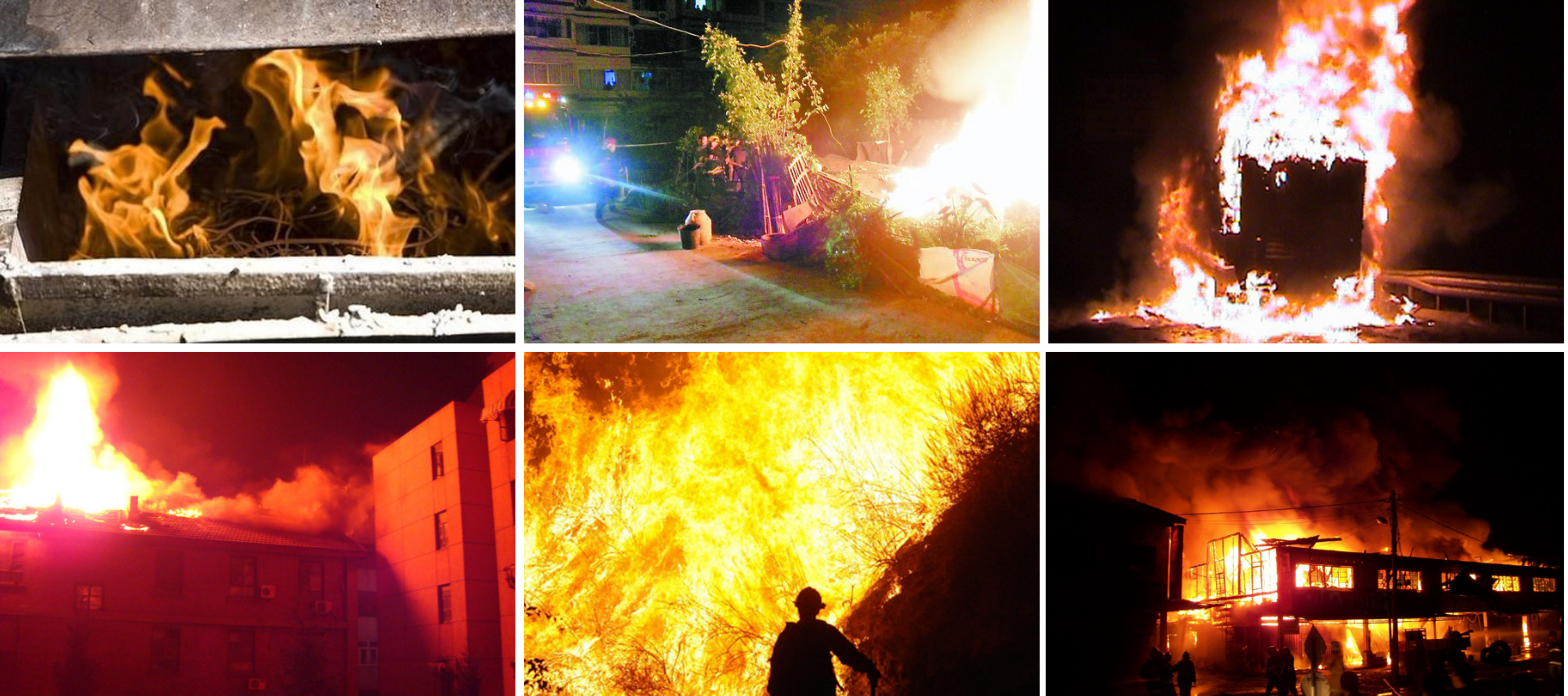}

\end{center}
\caption{Various appearances of fire.}
\label{fig:VariousFire}

\end{figure}

\begin{figure*}[htb]
\begin{center}
\includegraphics[width=0.9\linewidth]{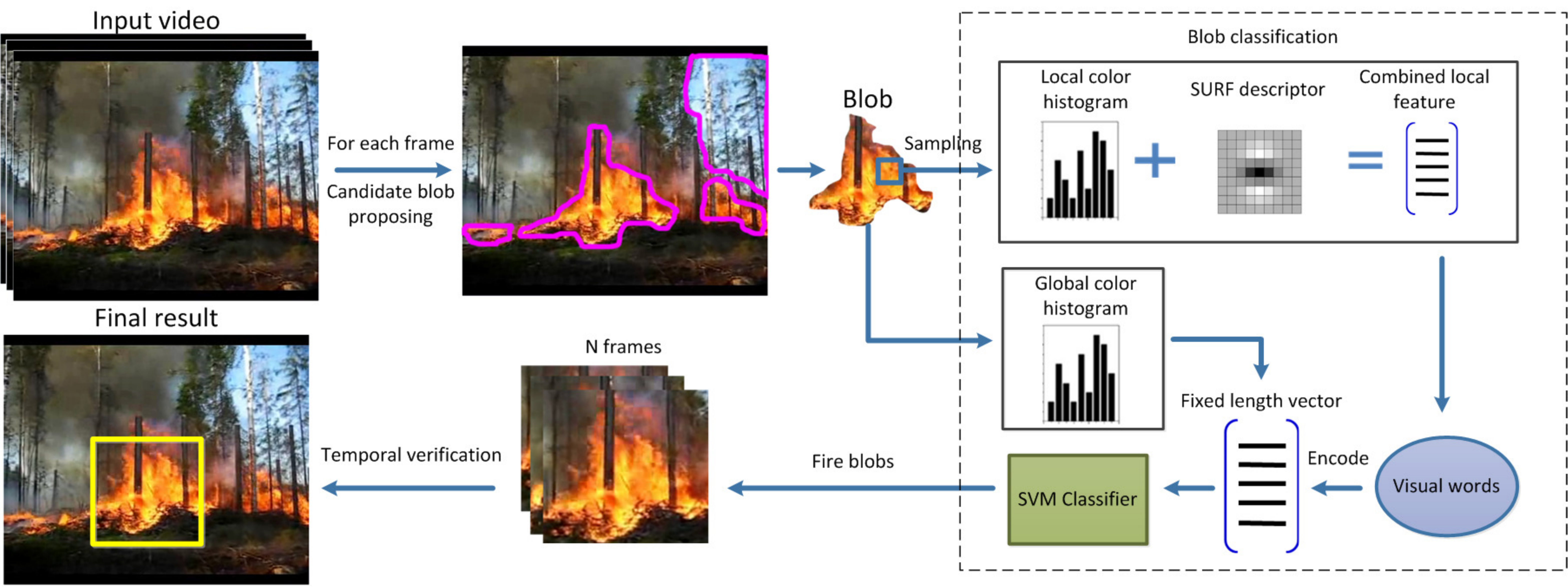}
\caption{Framework of proposed fire detection system.}
\end{center}

\label{fig:system_chart}
\end{figure*}

Existing fire detection methods can be classified into two categories: sensor-based methods and vision-based
methods. Sensor-based methods employed information captured by special instruments such as infra-red sensors and smoke sensors, which are expensive and not easy to get~\cite{Luo2007}. Moreover, those sensors typically detect the presence of certain particles generated by fire ionisation or photometry rather than the combustion itself. Therefore, fire-alarm systems based on those sensors usually have time delay and result in high false rates~\cite{Podraj08}. In contrast, vision-based methods used the normal R-G-B images or videos have several advantages~\cite{Healey93cvpr}\cite{Toreyin07CVPR}\cite{Ko09FireSVM}. Firstly, cameras are becoming more functional with steady dropping price. Thus images and videos can provide more detailed visual information for fire detection cheaply. Moreover, surveillance cameras already installed in public places can also be used for fire detection. Secondly, the response time can be faster than traditional sensors, as cameras do not need to wait for the smoke or heat to diffuse. Finally, compared to traditional point sensors, cameras can monitor a broader area, creating a higher possibility of fire detection at early stage~\cite{Ko09FireSVM}. Our system presented in this paper belongs to the latter category, and aims to detect fire in video clips.

Despite of the growing needs and interests in fire detection, there is still not a large number of work on fire detection in the computer vision literature~\cite{PVKBorges10}. Building a robust fire detection system is challenging in following two aspects: (1) fire or flame is flexible in shape and intensity, and it has no fixed structure or appearance. Though color is a relative distinct feature widely used for fire detection, the fire color appeared in images and videos is affected by the camera quality (e.g. resolution, sharpness) and settings (e.g. white balance). The fire also appears widely various in scale. Fig~\ref{fig:VariousFire} shows six fire images with different shape, intensity, color, and scale. Thus, it is hard to build a solid and generic model for fire. And (2) real application requires a fast fire detection method that can work in real-time. Complicated methods or models can not find a position in real applications. Some fire detection methods used filter banks~\cite{Toreyin07CVPR}, frequency transforms~\cite{Liu04ICPR}, and motion tracking techniques, thus need more computational processing time and not suitable for real-time application~\cite{PVKBorges10}.

There are mainly four kinds of limitations in existing fire detection systems: (1) merely based on color, or color plus motion information. Such as \cite{Healey93cvpr} and \cite{Celik07} only used color to detect fire. Limited clues may lead to high false alarm. (2) merely worked for some special situations, such as tunnel fire~\cite{Lee2007}. (3) used many heuristic fixed thresholds, such as \cite{Chen04ICIP}, \cite{Phillips02flame} and \cite{Celik07}, which restricted their applications. And (4) tested on limited data set, such as Cho et al.\cite{cho08} tested their method on six fire video clips, and Ko et al.\cite{Ko09FireSVM} performed their experiments on twelve video clips. Though they showed good results on the small data set, their systems may be impractical for other untested situation. Furthermore, as far as we know, there are no standard fire data sets.

Above mention four limitations lead to the big gap to real-world fire detection applications. The method we proposed in this paper is expected to narrow the gap. We have randomly downloaded 64 fire video clips from public video sharing web sites~\footnote{Project page. The data set and supplementary materials can be downloaded from the anonymous web page: \url{https://sites.google.com/site/firedetection2010/}}. This data set is much larger than those used by other researchers. We carried out a comprehensive empirical study and experiments on this data set. Based on the experiment results, we adopt three most effective fire features, including a local and a global color descriptor in CIE Lab space, and a SURF (Speeded Up Robust Features~\cite{Bay08surf}) texture descriptor. Then a Support Vector Machine (SVM) is employed to find fire regions. Based on our experiment, the RBF(Radial Basis Function) SVM kernel works the best for fire detection. Different with most existing fire detection methods which used only local features on pixel level, we find that our proposed global color feature and local SURF texture feature on regions made a good contribution for fire detection. Finally, a temporal verification is applied to reduce the false alarm. The flowchart of our system is shown in Fig.\ref{fig:system_chart}. Our system did not use any complicated features or tools, thus it is fast and can detect fire in real-time. Compared to Toreyin and Cetin's method proposed in~\cite{Toreyin07CVPR}, ours has higher recall and precision in fire detection on our large testing data set.
The remainder of this paper is organized as follows.

Section~\ref{sec:empiricalstudy} discusses the detailed implementation of our system based on empirical study.
Section~\ref{sec:system} introduces the framework of our fire detection system.
Section~\ref{sec:results} presents our experiment results. Comparisons with Toreyin and Cetin's method proposed in~\cite{Toreyin07CVPR} are also given in this section.
Finally Section~\ref{sec:summary} summarizes this paper and draws the conclusion.

\begin{figure*}[!ht]
\begin{center}
\includegraphics[width=1.0\linewidth]{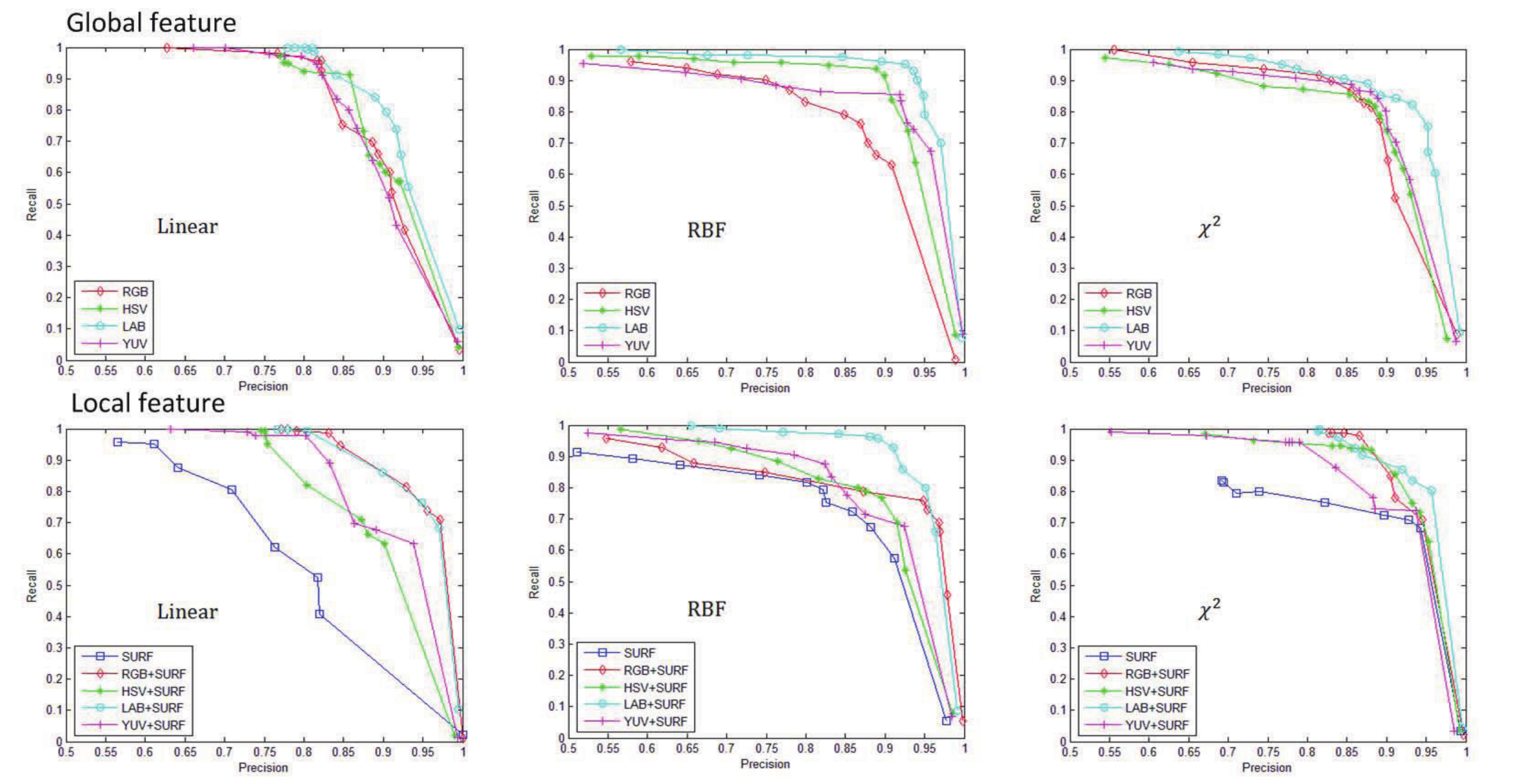}
\end{center}
   \caption{Precision-recall curves of different features with 3 kinds of SVM kernels.}
\label{fig:PRcurves}
\end{figure*}

\section{Empirical Study and Feature Selection}
\label{sec:empiricalstudy}

Our objective is to propose a robust real-time fire detection method for real-world application. However, as we discussed in section~\ref{sec:intro}, to build a solid and general model for fire is hard. Instead, we aim to implement the system based on a comprehensive empirical study. Since there are little open data set available for fire detection, we firstly collected a bundle of fire videos, and then carried out comprehensive experiments on it to figure out the empirically best feature and classifier kernel for fire detection.
According to our survey, neither theoretical analysis nor empirical comparisons on fire feature selection has been done before.

\subsection{Data set collection}
There are some small data set for fire detection available on internet, such as the data set published by Cetin et. al includes 13 video clips~\cite{VisiFire}. In order to build a larger data set, we downloaded the small public data sets and randomly downloaded fire video clips from public video sharing web sites, such as youtube.com and youku.com. We also captured some fire videos by ourselves to expand the data set.
Finally we have 64 video clips. The average length of each clip is 2 minutes. We tried to cover most of the application scenarios including indoor, urban outdoor, and forest fire with both static and moving background. Potential false alarms cases in the data set include car lights, human wearing fire-color clothes, illumination changing, and so on.

\subsection{Experiment settings}

Color and texture are the most widely used features for general object detection and recognition. Both features can be computed efficiently, hence suitable for real-time systems. Almost all fire detection methods use color as a distinct feature of fire, but explored it in different color space. Such as methods in \cite{Toreyin07CVPR}\cite{Phillips02flame}\cite{PVKBorges10} used RGB color information, and method in \cite{HOrng05} in used HSI color space. There are also other color spaces such as YUV and LAB. However, no trial on theoretical analysis and empirical comparison on color feature selection.

We capture key frames from videos and manually crop out image patches as samples for analysis. We collected 726 fire patches varied in color, shape, and intensity, and 637 non-fire samples involving a variety of objects including human and cars, which are full of visual details, and plain objects such as ceilings, walls, and skies, which are simple in color and texture. We randomly select 4/5 of the patches for training and 1/5 for testing.

We exploit Support Vector Machine (SVM) to do fire classification, since SVM is one of the latest and proved most successful statistical pattern classifier. The LIBSVM~\cite{CC01aLIBSVM} is used in our system. The weights of positive and negative sample are balanced with respect to their number. We search for optimized parameters using 5-fold cross validation with a parameter range of $2^{-8}$ to $2^8$. We test the system with three different SVM kernels: linear, RBF, and $\chi^2$~\cite{Koen10PAMI}.

\subsection{Global color feature}
We experiment with four color spaces including RGB, YUV, HSV, and LAB. The global color histogram of different color space consists of 96 bins, which is the concatenation of three 32-bin histograms for each channel. The precision-recall curves are shown on the left column in Fig.~\ref{fig:PRcurves}. As the result indicates, RGB histogram gives the poorest result, YUV, and HSV are better, and LAB performs the best in each SVM kernel setting. The highest performance is obtained by LAB histogram with RBF kernel.

\begin{figure}[htb]
\begin{center}
\includegraphics[width=0.85\linewidth]{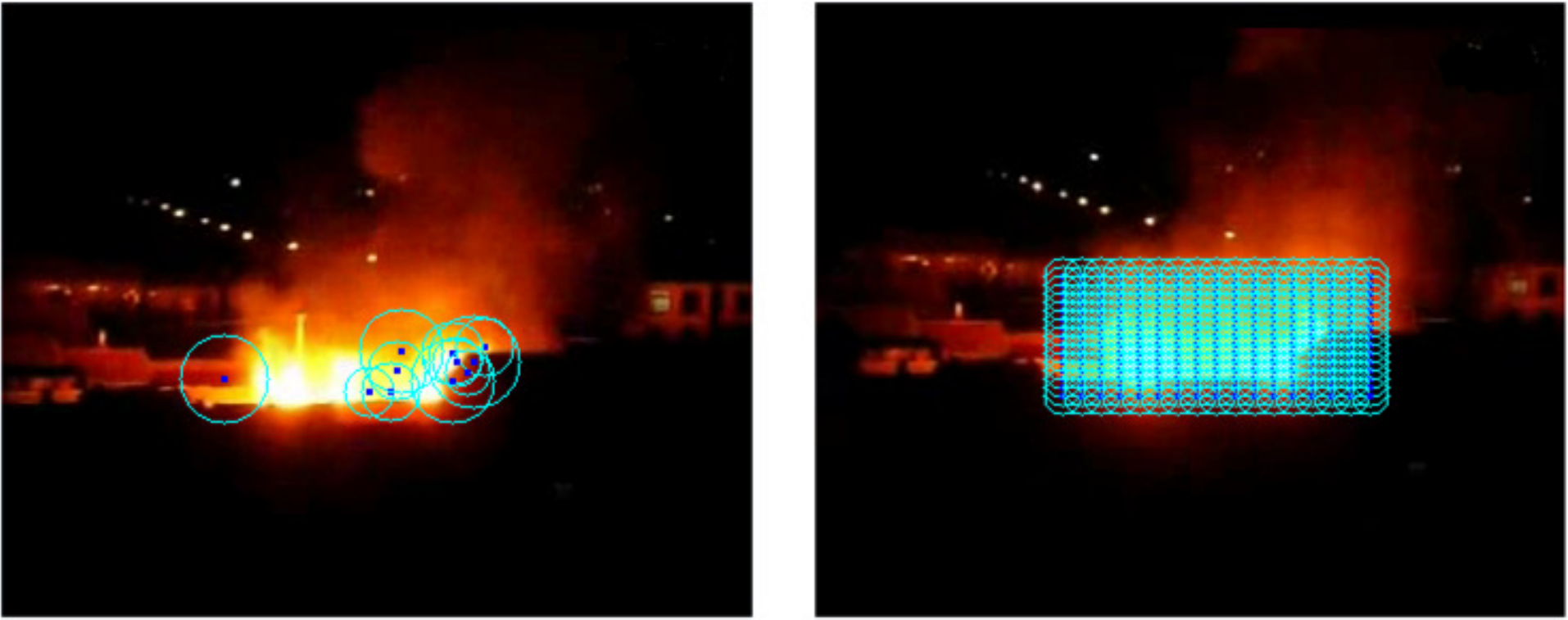}
\end{center}
   \caption{The left image illustrates keypoint sampling and the right image illustrates dense sampling. Blue dot is the location of sampling point and the diameter of cyan circle represents the size of SURF kernel.}
\label{fig:keypoint}
\end{figure}

\subsection{Local color-texture feature}

Texture is another distinctive feature besides color. Many texture descriptors like SIFT~\cite{Lowe99ICCV}, and HOG~\cite{Dalal05CVPR} have been used widely and successfully on various kinds of object and scene recognition. However, the high computational complexity limits their application in video surveillance and retrieval. To ensure that our system can run in real-time, we adopt the SURF descriptor~\cite{Bay08surf}. It has similar performance compared to SIFT but much faster to compute. Moreover, SURF computing is parallelizable, so that it can benefit from systems with multi-core CPUs and GPUs.

Actually, both SIFT and SURF operate on gray scale images only. There are several methods to boost color saliency of SIFT descriptor, like Hue-SIFT~\cite{VDW06PAMI} and CSIFT~\cite{Abdel06CVPR}. The former is a concatenation of hue histogram with SIFT descriptor, and the latter computes SIFT descriptor on R, G, and B channels respectively. Other methods and their performance discussed in~\cite{Koen10PAMI} have proved that local color information can improve the robustness and distinctiveness of SIFT descriptor by a good extent. Methods like CSIFT needs to compute 3 times over a single point and generates a descriptor with very high dimension, which slows down the matching process. In our method, we attach a local color histogram, which describes the color statistics within a SURF kernel, to a SURF descriptor.

The SURF descriptors are computed over the keypoints detected by its Fast Hessian Detector, yielding many 64 dimensional vectors. Because the quality and sharpness of video frames are relatively low, the threshold of the detector is set to 100 in order to detect more keypoints. The local color histograms are computed within the scope of each SURF kernel. Its size is 24 bin, with 8 bins for each channel. They are combined to form an 88-dimensional feature vector.

We need to obtain a fixed length vector representing the whole image, thus we used a visual vocabulary, which is also known as ``codebook'', to quantize all local features sampled from the image.

To construct the codebook, we first sample, generate and collect local features from all fire and non-fire training image patches. Then we use k-means clustering algorithm to produce 500 centers after 50 iterations. The centers are the elements in a codebook, since they are the most representative local features appear in our training data. The feature searching, matching and encoding steps are the same as that in our system.

The precision-recall curves of five different local features are shown on the right column in Fig.~\ref{fig:PRcurves}. As can be seen from the figure, including color information can greatly improve the system performance. Overall, LAB histogram + SURF + RBF kernel achieves the best performance.

\begin{figure*}[!ht]
\begin{center}
\includegraphics[width=1.0\linewidth]{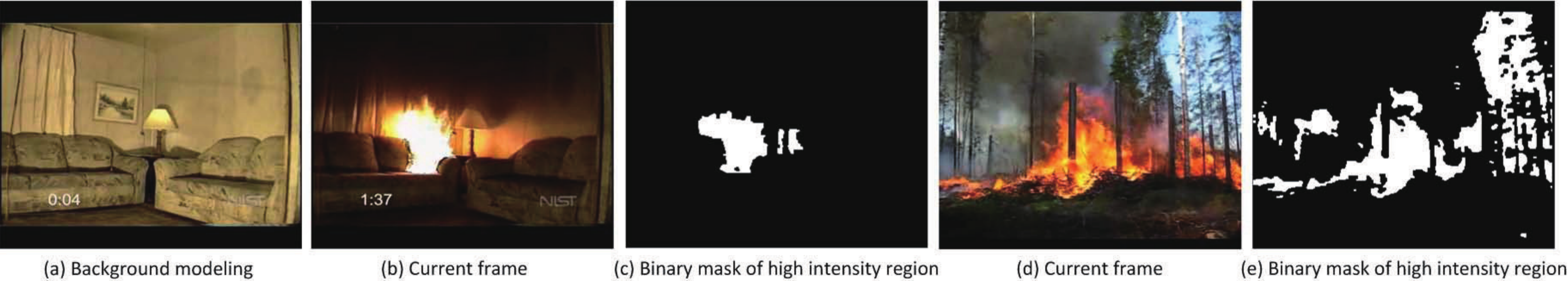}
\end{center}
   \caption{Two examples of candidate fire region proposing. The first row shows a surveillance video example with still background, the background modeling result is shown in (a), and the mask of candidate regions is shown in (c). The second row shows a video with moving background. The candidate fire region mask shown in (e) is found by a multi-level threshold.}
\label{fig:CandidateRegion}
\end{figure*}

\subsection{Keypoint sampling VS. dense sampling}
All previous experiments we conduct used keypoint sampling strategy in both codebook construction and classification. Recent researches like \cite{Jurie05ICCV} and \cite{NJT06} indicate that dense sampling gives better results than keypoint sampling in local feature based object classification. We now sample over the whole image using SURF kernels with several predetermined intervals and scales, in both codebook construction and classification stage. The difference between keypoint sampling and dense sampling is shown in Fig.~\ref{fig:keypoint}.

\begin{table}[htbp]
\centering
 \caption{Precision comparisons of two sampling strategies}
 \begin{tabular}{|c|c|c| }
   \hline\hline
    & dense sampling & keypoint sampling\\ \hline
   SURF & \textbf{84.8485} & 76.431 \\ \hline
   RGB + SURF & \textbf{86.1953} & 79.1245 \\ \hline
   HSV + SURF & 84.1751 & \textbf{ 85.1064} \\ \hline
   LAB + SURF & \textbf{92.2559} &  84.8484 \\ \hline
   YUV + SURF & \textbf{85.1852} & 80.4713 \\ \hline
\hline
    \end{tabular}
\label{tbl:T1}
\end{table}

Result of comparison between both sampling strategies is shown in Table.~\ref{tbl:T1} . We find that dense sampling outperforms keypoint sampling in most cases,which is consistent with the conclusion in~\cite{Jurie05ICCV}.  In smooth surfaces like walls and ceilings, Fast Hessian detector may find few keypoints, which is an impact on classification results. In contrast, dense sampling can gather sufficient information of an image, hence it is more robust.

\begin{table}[htbp]
\centering
\caption{ Accuracy comparison of SURF and SURF-128 }
 \begin{tabular}{|c|c|c| }
  \hline\hline
  & 64-dimension & 128-dimension\\ \hline
   SURF & 84.8485 & \textbf{85.5218} \\ \hline
   RGB + SURF & \textbf{86.1953} & 85.5218 \\ \hline
   HSV + SURF & 84.1751 & \textbf{86.8687} \\ \hline
   LAB + SURF &  92.2559 &  \textbf{92.5925} \\ \hline
   YUV + SURF & 85.1852 & \textbf{86.532} \\ \hline
\hline
    \end{tabular}
\label{tbl:T2}
\end{table}

\subsection{SURF VS. SURF-128}
SURF-128 is an extended SURF descriptor that has 128 dimensions. It calculates the sums of positive and negative Laplacian separately, and the resulting signs in the descriptor distinguish positive gradient change from negative one. In our problem, it may be able to distinguish bright flame from dark background. Although SURF-128 is generally more distinctive than standard 64-dimensional SURF, it will significantly slow down the matching step. We compare SURF-128 with standard SURF to find out whether it deserves the extra cost. As shown in Table~\ref{tbl:T2}, on average, SURF-128 is only slightly better. In the worst case, SURF-128 is worse than standard SURF. Therefore, we abandon SURF-128 for our real-time fire detection.

\section{System Implementation}\label{sec:system}
As shown in Fig.\ref{fig:system_chart}, our fire detection system consists of three cascaded parts: (1) candidate fire regions proposing by a background model, (2) fire region classifying with color-texture features and a dictionary of visual words, and (3) temporal verifying. We will discuss each part in detail in following three sections respectively.

\subsection{Candidate fire region proposing}
Most of fire detection systems detect fire regions or pixels directly. Based on observations and statistic experiments, we find that fire regions are relatively bright, i.e. their pixel intensities are above some certain value. We can use this feature to find fire region candidates and fasten the further fire detection process. According to our survey, only Toreyin et al.\cite{Toreyin07CVPR} tried to eliminate the non-fire background. However, they used a fix heuristic threshold, which may fail when the background is bright or when there are fire reflections on white wall. To deal with such challenging circumstances, we adopt a multi-level threshold based on empirical study. The system can adjust the threshold automatically according to the background intensity statistics. Higher background intensity leads to lower threshold, and vice versa.

For a surveillance video clip, which is usually captured by a fixed camera, the background will be relatively static. We applied background substraction to further reduce the fire region candidates. Two examples are shown in Fig.~\ref{fig:CandidateRegion}. The three images on the first row illustrate a fire example taken by a surveillance camera. The first image shows the result of background modeling, and the right most white mask shows the proposed candidate fire regions. Two images in the second row illustrate an example fire video with moving background. We can not model the background easily, so we use the multi-level threshold to find the fire region candidates, as shown in the image on bottom right.

\subsection{Feature extraction and region classification}

At the initialization stage of classification, we use the pre-generated codebook to build a K-D tree, which allows fast feature indexing and searching. For each incoming blob, we densely sample LAB+SURF descriptors from its region with interval of 9 pixels and SURF kernels of $9\times9$ scale. This step produces $N$ descriptors.

For each descriptor $d_j$ ,$j = 1, 2, ..., N$, we query its distance $D(d_j, v_i)$ to $m$ nearest neighbors ${v_i|i = 1, 2, ..., m}$ from the K-D tree. In order to prevent drawbacks of single matching, $d_j$ can be matched with all m nearest neighbors. Theoretically, the larger m is set, the better classification result will get, and the higher computational cost is needed. According to our experiments, $m=10$ can achieve the best cost/performance ratio. The weight of matching $d_j$ with $v_i$ is calculated as following:
\begin{equation} \label{eq:weight}
w_{ij} = \frac{K_\sigma(D(d_j,v_i))}{\sum^{m}_{k=1}K_\sigma(D(d_j,v_k))}, i\in[1, m]
\end{equation}

Where $K_\sigma(x)$  is a Gaussian kernel function defined as:

\begin{equation} \label{eq:Gkernel}
K_\sigma(x) = \frac{1}{\sqrt{2\pi\sigma}}exp(-\frac{1}{2}\frac{x^2}{\sigma^2})
\end{equation}

While using this matching method, we assume that the distance between $d_j$ to $v_i$ satisfies Gaussian distribution due to the effect of clustering. This kernel converts smaller distance into higher probability.

We use a 500-bin histogram $h$ to encode all local features in a fire candidate blob. The values of each bin is calculated as following:
\begin{equation} \label{eq:bin500}
h_i = \sum_{j=1}^{N}w_{ij}, i\in[1, 500]
\end{equation}

After normalization, each bin represents the probability of occurrence of the correspondent element in the codebook and the histogram.
Finally, we compute the 96-bin normalized global LAB histogram for the fire candidate blob. The concatenation of both histograms is the final representation of the blob, and is used as the input to the SVM classifier.

\subsection{Temporal verification}
Objects like street lamps and car lights have great similarity with fire in appearance, and they are hard to be excluded based on a single frame. Thus, blobs classified as fire in the previous step should be further verified using statistic of temporal variation.

\begin{figure}[htb]
\begin{center}
\includegraphics[width=0.98\linewidth]{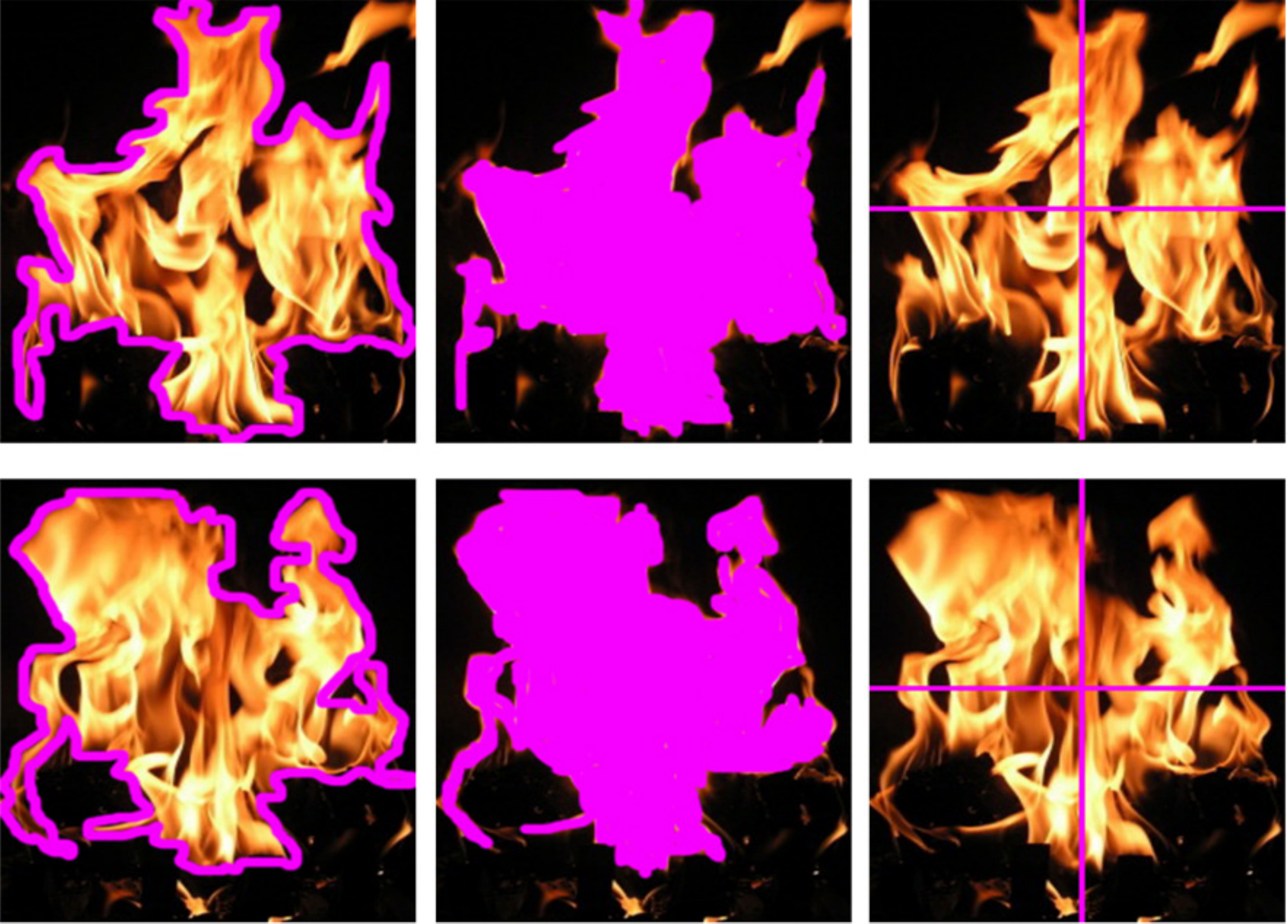}
\end{center}
   \caption{The first row and the second row are two different frame of the same video respectively. From left to right: illustration of perimeter, area and spatial distribution.}
\label{fig:7}
\end{figure}

Fire mainly exhibit shape variation but not color or texture. When a new blob emerges, we use three simple parameters: perimeter, area, and spatial distribution to estimate the stableness of the region over consecutive 25 frames, without having to figure out its exact shape representation like Fourier descriptors used in~\cite{Liu04ICPR}. Perimeter and area are very intuitive, and $\mu_p, \mu_a, \sigma_p, \sigma_a$ denote their mean and standard deviation respectively. To calculate spatial distribution, we divide the bounding rectangle of a blob into 4 equal sub rectangles, and calculate the number of pixels within each sub rectangle that are labeled as ``$1$'' in binary mask, namely $d_1, d_2, d_3, d_4$. The standard deviation of spatial distribution is defined as follow:
\begin{equation} \label{eq:sigma}
\sigma_d = \sigma_{d_1} + \sigma_{d_2} + \sigma_{d_3} + \sigma_{d_4}
\end{equation}

We define that a blob is stable if the parameters satisfy the following condition:
\begin{equation} \label{eq:cond1}
\sigma_p < t_1\mu_p ~~~~ and ~~~~ \sigma_a < t_1\mu_a ~~~~ and ~~~~ \sigma_d < t_1\mu_a
\end{equation}

And a blob is unstable if the following condition holds:
\begin{equation} \label{eq:cond2}
\sigma_p > t_2\mu_p ~~~~or~~~~ \sigma_a < t_2\mu_a ~~~~or~~~~ \sigma_d > t_2\mu_a
\end{equation}

Where $t_1$ and $t_2$ are thresholds based on environmental settings. Indoor surveillance can use lower values. In outdoor scenarios, where fire may be greatly influenced by air flow, the thresholds are higher. False alarm rate can be further pruned by these two conditions, i.e. Satisfying Equation(\ref{eq:cond1}) and (\ref{eq:cond2}) should not be proposed as fire.  For example, if we use motion tracking and finds that a newly emerged blob vanishes from its original location after several frames. We can assert that it is not fire because fire does not move by itself. There may be other prior knowledge which we can take advantage of, but they are application specific and out of scope of this paper.

\section{Results and Discussion}\label{sec:results}

We use our newly collected dataset to benchmark the performance of our method, and we also test Toreyin et al.'s method~\cite{Toreyin07CVPR}  for comparison. Since both methods make a decision at an interval of several frames, we do not use a frame-by-frame evaluation. We manually split the videos into individual sections of 200 frames, and label them as either ``contains fire'' or ``does not contain fire''. As a result, we have 744 sections as testing units and we are able to benchmark using precision and recall rate. The result is shown in Table.~\ref{tbl:resultcompare1}.

Overall, our method has similar precision rate comparing to Toreyin's method. However, we significantly outperform Toreyin's method in recall rate. We should be aware that miss in fire detection is much more critical than false alarm. Therefore, our method is more preferable for real-life applications.

We also compared our method and Toreyin's method for typical instances picked out from the data set, as shown in Table.~\ref{tbl:resultcompare2}. The typical scenarios are described briefly in the right column as video content description. We can see our result is much better than Toreyin's method. However, when inspecting the result, we find some typical failure cases. Both our method and  Toreyin's fail to detect the fire on initial combustion, because the size of the flame is very small, both methods are not sensitive enough for it. For night traffic videos, both methods issue some false alarms on car lights, since the car light is in bright yellow tint and moving in video, which is similar to the characteristics of fire .

Due to the na\"{\i}ve thresholding in pixel color, Toreyin's method always fail to recognize fire with bright intensity, which is the main reason of its low recall rate. Our method, however, overcomes this problem and gains great advantage. Toreyin's method also issues false alarms on human skin and yellow shirts. In contrast, our method withstands the challenge.

\begin{table}
\centering
    \caption{Overall benchmark performance.}
    \begin{tabular}{|c|c|c|}
    \hline
                    & Our method & Toreyin et al.~\cite{Toreyin07CVPR}   \\ \hline
    True positive   & 361        & 311       \\ \hline
    True negative   & 305        & 308       \\ \hline
    False positive  & 27         & 24        \\ \hline
    False negative  & 81         & 131       \\ \hline
    Precision rate  & 93.04\%     & 92.83\%    \\ \hline
    Recall rate     & 81.67\%     & 70.36\%    \\ \hline
    \end{tabular}
    \label{tbl:resultcompare1}
\end{table}

\newcommand{\tabincell}[2]{\begin{tabular}{@{}#1@{}}#2\end{tabular}}

\begin{table*}
\centering
    \caption{Detection results for typical instances from the data set.}
    \begin{tabular}{ |c|m{1.2cm}<{\centering}|m{1.2cm}<{\centering}|m{1.2cm}<{\centering}|m{1.2cm}<{\centering}|m{3.5cm}<{\centering}|}
    \hline
    \ & \multicolumn{2}{|c|}{Our method} & \multicolumn{2}{|c|}{ Toreyin et al.~\cite{Toreyin07CVPR} } &   	\\    \hline
    Input    & \small{False positive} & \small{False negative} & \small{False positive} & \small{False negative} & \small{Video Content Description} \\ \hline
    Video 1\tabincell{c}  & 0 		    & 0              & 0              & 6         & \small{A Christmas tree is burning in a bright room} \\ \hline
    Video 2  & 4              & 0              & 9              & 1         & \small{Flash lamp and dazzling flame in a warehouse} \\ \hline
    Video 3  & 0              & 0              & 0              & 9         & \small{A man doing fire experiment in a dark place} \\ \hline
    Video 4  & 0              & 0              & 0              & 5         & \small{White flame burning in a back yard} \\ \hline
    Video 5  & 0              & 0              & 1              & 0         & \small{A man wearing an orange shirt is walking around} \\ \hline
    Video 6  & 0              & 0              & 4              & 0         & \small{Bank surveillance video of human face} \\ \hline
    \end{tabular}
    \label{tbl:resultcompare2}
\end{table*}


\section{Summary} \label{sec:summary}
In this paper, we present a robust real-time fire detection method based on empirical study. To carry out the empirical experiments, we collected so far the largest open data set for fire detection in video. Experiments show that, overall, LAB histogram plus SURF texture descriptor, plus the RBF SVM kernel, lead to the best performance of an 82\% recall with 93\% precision on the data set, which greatly improved the performance by state-of-the-arts methods.

For future works, we can study more detailed fire characteristics to effectively detect the fire on initial combustion and exclude the fire-like objects such as car light in night video.


\bibliographystyle{spmpsci}      
\bibliography{fire2010}


%
%


\end{document}